# Non-monotonic Reasoning and the Reversibility of Belief Change

Daniel Hunter
3131 E. Highway 246
Santa Ynez, CA 93460

## Abstract

Traditional approaches to non-monotonic reasoning fail to satisfy a number of plausible axioms for belief revision and suffer from conceptual difficulties as well. Recent work on *ranked preferential models* (RPMs) promises to overcome some of these difficulties. Here we show that RPMs are not adequate to handle *iterated* belief change. Specifically, we show that RPMs do not always allow for the reversibility of belief change. This result indicates the need for numerical strengths of belief.

## 1 INTRODUCTION

Makinson (1989) and Kraus, Lehmann, and Magidor (1990) give axioms for a relation of non-monotonic inference and show that most well-known systems for non-monotonic reasoning -- default logic, circumscription, McDermott and Doyle's modal systems -- fail to satisfy one or more of these axioms. Hanks and McDermott (1987) describe anomalies in the application of default rules and circumscription to an intuitive case of non-monotonic reasoning. In general, there is a growing awareness of the inadequacy of traditional approaches to non-monotonic reasoning.

Recent work on preferential and ranked preferential models (Kraus, Lehmann, and Magidor, 1990; Lehmann, 1989; Makinson, 1989) overcomes some of these difficulties. In a preferential model, the worlds or states are related by a binary preference relation $<$. An inference relation $\mathrel{\vdash}$ is defined by saying that $A \mathrel{\vdash} B$ iff $B$ is true in all the most preferred worlds in which $A$ is true. Depending on the properties of $<$, various non-monotonic logics result from this definition. In a ranked preferential model (RPM), the preference relation may be thought of as stemming from a well-ordering of some partition of the worlds (so that where $w$ and $v$ are worlds, $w < v$ iff $w$ occurs in a partition element preceding the partition element to which $v$ belongs). Thus the worlds are in essence *ranked*, with ties permitted.

The purpose of this paper is to show that despite their advantages over other approaches to non-monotonic reasoning, RPMs are still inadequate. In particular, they cannot adequately handle *iterated* belief change, a point already made by Spohn (1988). Here we give a formal proof of this claim. The proof shows that RPMs cannot handle *reversibility* of belief change: sometimes we learn that a piece of information we thought true is not true and we wish to revise our beliefs by going back to the state of belief we had before the information was given.

We first discuss the relation between non-monotonic reasoning and belief change. Next, we demonstrate the inadequacy of RPMs to handle the reversibility of belief change. Finally, we examine one possible way of remedying the deficiency and conclude that it fails.

## 2 BELIEF CHANGE AND INFERENCE

A *monotonic* inference relation $\vdash$ is one for which the following condition holds:

(M) If $A \vdash C$, then $A \& B \vdash C$.

A *non-monotonic* inference relation is one for which (M) sometimes fails. Common-sense reasoning is generally thought to be non-monotonic, as exemplified by the following ubiquitous example: on learning that Tweety is a bird, I leap to the conclusion that Tweety flies, but on next learning that Tweety is a penguin, I withdraw that conclusion. So in ordinary, or common-sense, reasoning, gaining additional information may cause previous inferences to be withdrawn.

A fruitful way of viewing common-sense reasoning of the non-monotonic variety is as a case of belief revision: one is willing to non-monotonically infer $B$ from *A* if adding *A* to one's stock of beliefs results in *B*'s being believed. This way of viewing non-monotonic inference suggests an investigation of the rules for rational belief revision, rather than seeking a weakening or modification of the classical relation of logical implication, which concerns static relations between propositions.

Gärdenfors (1988) presents a widely accepted set of axioms for belief revision. Let **B** stand for a person's set of beliefs at a particular time, which set is assumed to be deductively closed (we call such a deductively closed set a *belief set*). We let $\mathbf{B}_A^*$ stand for the belief set that results when a person with belief set **B** comes to believe proposition A -- i.e. it is the result of revising the beliefs in **B** to accommodate belief in A. $\mathbf{B}_A^*$ should be distinguished from what we shall denote as $\mathbf{B}_A^+$, the deductive closure of $\mathbf{B} \cup \{A\}$. The latter contains every proposition in **B**, the former need not if A contradicts the beliefs in **B**. In particular, if **B** contains $\neg A$, the negation of A, $\mathbf{B}_A^*$ will not contain $\neg A$ unless A is contradictory. Gärdenfors' axioms are the following:

(B1) $\mathbf{B}_A^*$ is a belief set.

(B2) $A \in \mathbf{B}_A^*$.

(B3) $\mathbf{B}_A^* \subseteq \mathbf{B}_A^+$.

(B4) If $\neg A \notin \mathbf{B}$, then $\mathbf{B}_A^+ \subseteq \mathbf{B}_A^*$.

(B5) $\mathbf{B}_A^*$ is inconsistent iff $\vdash \neg A$.

(B6) If $\vdash A \leftrightarrow B$, then $\mathbf{B}_A^* = \mathbf{B}_B^*$.

(B7) $\mathbf{B}_{A \& B}^* \subseteq (\mathbf{B}_A^*)_B^+$.

(B8) If $\neg B \notin \mathbf{B}_A^*$, then $(\mathbf{B}_A^*)_B^+ \subseteq \mathbf{B}_{A \& B}^*$.

(B1)-(B8) are equivalent to the finitary rules for what Lehmann (1989) calls *rational* inference.

Lehmann makes uses of a finitary inference relation $\vdash$, which is taken to be a relation of non-monotonic implication between formulas. Here we regard it as a relation between nonlinguistic propositions. The rules listed in (Lehmann 1989) for $\vdash$ translate into Gärdenfors' terminology of belief set revision by equating $A \vdash B$ with $B \in \mathbf{B}_A^*$. The next section discusses models of belief change satisfying the above axioms.

## 3 SEMANTICS FOR BELIEF CHANGE

Ranked preferential models provide a semantics for rational belief change. Here we apply RPMs within the general possible worlds framework, taking possible worlds as non-linguistic entities relative to which propositions are true or false. As usual, we identify a proposition with the set of worlds in which it is true and we assume that every set of worlds corresponds to a proposition.

Let W be the set of all possible worlds. Within this framework, an RPM may be considered a well-ordered partition of W, that is, a sequence $E_0, E_1, \ldots$ of disjoint subsets of W such that $\bigcup_{i=1}^{\infty} E_i = W$[1]. A well-ordered partition of worlds represents a state of belief in the following sense. Worlds within the same partition element are equally equally believable or disbelievable; Worlds in a given partition element are more believable than worlds occurring in succeeding partition elements. The members of the initial partition element $E_0$ are all the worlds that are not disbelieved -- i.e. no world in $E_0$ is believed not to hold. The worlds in the remaining partition elements are disbelieved.

A well-ordered partition yields a belief set in the following manner. Recall that a belief set is just a deductively closed set of propositions, representing the beliefs of some agent. The initial element of the well-ordered partition, $E_0$, is the *total content* of the agent's belief. It can be thought of as the (possibly infinite) conjunction of all the propositions believed by the agent. A proposition is in the belief set of an agent iff it is entailed by the total content of the agent's belief. Within the set-theoretic representation of propositions with which we are working, proposition A entails proposition B iff $A \subseteq B$. Thus the belief set of the agent can be defined to be $\{A \subseteq W : E_0 \subseteq A\}$.

---

[1] We do not distinguish, as does Lehmann (1989), between states and worlds. Given our assumption that the objects of belief and inference are propositions, not sentences, and that a proposition is any subset of worlds, the smoothness condition and the condition that the ranking of worlds be derived from a total order (Lehmann, 1989, p. 215) together imply that the ranking of worlds constitutes a well-ordered partition.

An RPM determines how beliefs are to be revised when new information is received. Suppose proposition A comes to be believed. The task is to say what the new total content of belief is. The rule is this: Let $E_i$ be the first partition element whose intersection with A is non-empty (i.e. the first partition element consistent with A). Then the new total content of belief is $E_i \cap A$. Hence proposition B will be believed in the result of revising the agent's beliefs to accommodate A iff $E_i \cap A \subseteq B$.

## 4 ITERATED BELIEF CHANGE AND REVERSIBILITY

Wolfgang Spohn (1988) criticized the theory of belief change just presented on the grounds that it cannot account for iterated belief change. To handle iterated belief change, argued Spohn, one must know what the new ranking of worlds is after a belief change. RPM semantics only tells us what is believed after a single belief change; it does not tell us how the ranking of worlds changes. The reason this is a problem is that in general the result of a belief change depends, not just upon what is believed, but also upon epistemic preferences among disbelieved propositions. If all we know after a belief change is what propositions are believed, but not what the new ranking of worlds is, there is no way in general to determine the result of future belief changes.

Spohn's solution was to assign numerical degrees of disbelief to worlds together with a rule for revising those degrees of disbelief when new information is obtained. (For details see Spohn, 1988). Given the strong motivation in the non-monotonic reasoning literature to avoid numerical approaches, this solution may appear unsatisfactory to many. Might it not be possible to supplement the theory of ranked preferential models with a rule for revising the ranking of worlds when a belief change occurs? Spohn considered two ways of doing so and showed that both fail. He concluded that accounting for iterated belief change in terms of well-ordered partitions "... looks hopeless" (1988, p. 114). But is it really hopeless? Perhaps there is an acceptable way of revising rankings that Spohn overlooked. The next task is to show that Spohn was correct in his pessimism.

A theorem proved by Lehmann and Magidor (Kraus, Lehmann, and Magidor, 1990, p. 216) appears at first glance to show Spohn wrong. Formulated in terms of belief sets, the theorem says that a belief revision function satisfies (B1)-(B8) if and only if it is defined by some RPM. This result seems to imply that RPMs capture exactly the logic of belief change.

The resolution of this difficulty is that while (B1)-(B8) may completely capture the logic of a single step of belief change, they do not completely capture the logic of *iterated* belief change. To distinguish between RPMs and Spohnian belief revision, additional axioms for iterated belief change are needed. Two axioms for iterated belief change are proposed below, but there is no claim that these axioms are complete.

The first axiom concerns what happens when more precise information is obtained. Suppose the agent comes to believe proposition A and revises her beliefs accordingly. Suppose next that the agent comes to believe B, where B entails A. That is, the agent gets more precise information. What should the agent's beliefs be after receiving the second piece of information? I want to say that her beliefs should be exactly the beliefs she would have had if only the second piece of information had been received. Formally, this amounts to the axiom:

(B9) If $B \vdash A$, then $(B^*_A)^*_B = B^*_B$.

Thus according to (B9), if the agent first comes to believe that some object is, say, a tree and then comes to believe that it is a pine tree, her beliefs should be the same as they would have been had she initially come to believe it was a pine tree.

The second axiom is analogous, this time dealing with conflicting pieces of information. If the agent first comes to believe A and next comes to believe B, where A and B are inconsistent, I want to say that the net effect of these two changes is just as if only the latter had occurred. That is, coming to accept a belief that conflicts with a previous piece of information "wipes out" the effect of the previous information. Thus the axiom:

(B10) If $B \vdash \neg A$, then $(B^*_A)^*_B = B^*_B$.

Spohn's system of belief revision satisfies both (B9) and (B10).

To show that RPMs are inadequate to handle iterated belief change, assume that some rule for revising rankings of worlds is given for RPMs. (B9) and (B10) imply the following regarding any such rule: that when a belief change involves just the proposition A, the *relative* rankings of worlds within the sets A, $\neg A$ remain the same. That is, if worlds $w_1$ and $w_2$ are both in A and $w_1$ precedes $w_2$, then after updating on A, $w_1$ still precedes $w_2$. Similarly for worlds within $\neg A$. To show this, let B be the current belief set, < the precedence relation between worlds, and note that by the updating rule for RPMs the following conditions are equivalent for worlds $w_1, w_2$:

$$w_1 < w_2.$$

and

$$B^*_{\{w_1,w_2\}} = \{w_1\}.$$

Let $<$ be the old precedence relation and let $<'$ be the new precedence relation determined by revising the old ranking to accommodate belief in proposition A.

Suppose $w_1, w_2 \in A$ and $w_1 < w_2$. By (B9),

$$(B^*_A)^*_{\{w_1,w_2\}} = B^*_{\{w_1,w_2\}} \text{ and since } w_1 < w_2,$$

$B^*_{\{w_1,w_2\}} = \{w_1\}$. Hence $(B^*_A)^*_{\{w_1,w_2\}} = \{w_1\}$, so $w_1 <' w_2$. A similar argument can be given for the case in which $w_1, w_2$ belong to $\neg A$, by appealing to (B10).

To show that RPMs do not allow for the reversibility of belief change in all cases, we must be more precise about what reversibility amounts to. In abstract form, a theory of belief revision is a function $f$ from the cross-product space of belief states and epistemic inputs to the space of belief states. Reversibility means that for any belief state $S$ and epistemic input $E$, there is an epistemic input $E'$ such that $S = f(f(S,E),E')$. What we learn, we can unlearn.

Should any conditions be placed on the epistemic input that returns us to the previous state? It is reasonable, I think, to require that it only involve the proposition that caused the belief change in the first place. Otherwise, there is too much leeway: we could cheat and use information about the starting belief state to pick the right proposition or sequence of propositions to get back to where we started. More formally, let an epistemic input be an ordered pair $<A,\alpha>$, where A is a proposition and $\alpha$ is an *epistemic attitude*. In the Spohn system, for example, an epistemic attitude is a strength of belief, so the epistemic input $<A,\alpha>$ represents coming to believe coming to believe A with strength $\alpha$. Then the reversibility condition says:

(R) If $S$ is a belief state, $A$ a proposition, and $\alpha$ an epistemic attitude, then there exists an epistemic attitude $\beta$ such that $S = f(f(S,<A,\alpha>),<A,\beta>)$.

In the case of RPMs, an epistemic input is formally simply a proposition, but it is implicitly understood to be a proposition together with an attitude of belief towards that proposition. More generally, we will say that an epistemic input for an RPM is a proposition together with an attitude of belief, disbelief, or suspension of judgment (neither believing nor disbelieving). To allow numerical degrees of belief as inputs to RPMs would go against the spirit of much research in non-monotonic reasoning, but we will nonetheless later consider the possibility of deriving numerical degrees of belief from RPMs. For the time being, however, we assume that an epistemic input to a RPM is a non-numerical one of the sort just described.

Within the formalism of RPMs, an attitude of belief can be represented as the addition of a proposition (to the stock of beliefs) and the attitude of disbelief also by the addition of a proposition, namely the negation of the proposition disbelieved. A technical problem arises, though, for the attitude of suspension of belief. There seems to be no way to represent coming to suspend judgment in a proposition within the formalism of RPMs. This problem can be skirted if we define the belief set that results from suspending judgment in A as the intersection of the belief set resulting from belief in A with the belief set resulting from belief in $\neg A$. (That is, what you believe when you suspend judgment in a proposition is exactly what you would believe whether you believed or disbelieved the proposition in question.)

We are now in a position to argue that no belief revision rule for RPMs that satisfies (B9) and (B10) can also satisfy (R). First consider a simple model in which there are only four worlds, the four boolean atoms formed from the atomic propositions A and B. Thus the set of worlds W is $\{A\&B, A\&\neg B, \neg A\&B, \neg A\&\neg B\}$. Let RPM $r_1$ rank the worlds thus:

$\{\neg A\&B\}, \{A\&B, A\&\neg B, \neg A\&\neg B\}$ (the most believable worlds listed first). Let RPM $r_2$ rank the worlds so: $\{\neg A\&B\}, \{A\&B, A\&\neg B\}, \{\neg A\&\neg B\}$. If the revision rule satisfies both (B9) and (B10), then coming to believe A will take both $r_1$ and $r_2$ to the RPM $r_3$ = $\{A\&B, A\&\neg B\}, \{\neg A\&B\}, \{\neg A\&\neg B\}$. Are there epistemic inputs involving only proposition A that take $r_3$ into $r_1$ and into $r_2$? Neither belief in A nor suspension of judgment regarding A will take $r_3$ into either $r_1$ or $r_2$. Only disbelief in A -- i.e. belief in $\neg A$ -- will do so. But belief in $\neg A$ cannot simultaneously take $r_3$ to both $r_1$ and $r_2$. Without loss of generality, suppose belief in $\neg A$ takes $r_3$ into $r_2$. Then there is no way that $r_1$ can be recovered -- belief change is irreversible.

This argument can be generalized. Let W be any set of worlds, finite or infinite, with cardinality greater than three. Partition W into four non-empty subsets $W_1$, $W_2, W_3$, and $W_4$. Define A to be $W_1 \cup W_2$. Then coming to believe A will take both the rankings $<W_3$, $\neg W_3>$ and $<W_3, A, W_4>$ into the ranking $<A, W_3, W_4>$ and the argument goes through as before.

## 4.1 USING NUMERICAL STRENGTHS OF BELIEF

It might be suggested that the comparison between the Spohn system and RPMs is unfair because the former makes use of strengths of belief, but such epistemic resources are denied to the latter. Perhaps if the epistemic inputs to RPMs were expressed in terms of strengths of belief, reversibility of belief change could be achieved. This suggestion will not work, however, for the following reason. Consider the equivalent problem of defining degrees of *disbelief* in terms of RPMs. Let $d()$ be a disbelief function over the set of propositions. To derive $d()$ from a given RPM and to maintain consistency with the belief updating rule for RPMs, we must impose the following conditions on $d()$:

(i) For $w \in W$, $d(w)$ is a function of the rank of $w$.

(ii) For $A, B \subseteq W$, $d(A) < d(B)$ iff where $E_i$ is the first partition element consistent with $A \cup B$, $E_i \cap B = \emptyset$.

The first condition is imposed to capture the idea that knowing the rank of a world determines its degree of disbelief. The second condition is imposed to maintain consistency with the updating rule for RPMs: if the degree of disbelief in A is less than the degree of disbelief in B, then if $A \vee B$ becomes believed, A should be believed but B should still have some positive degree of disbelief; conversely, if coming to believe $A \vee B$ results in a belief state in which A is believed but B is disbelieved, then the initial degree of disbelief in A must have been less than the initial degree of disbelief in B.

These conditions imply that where A is a proposition, $d(A)$ must equal the minimal rank of the worlds in A. (i) and (ii) imply that $d()$'s range is isomorphic with the set of possible ranks; hence we may as well identify degrees of disbelief with the natural numbers $0, 1, ...,$ and identify the degree of disbelief in a world with that world's rank. Let A be a proposition and $E_A$ the first partition element consistent with A. Let $w^* \in E_A \cap A$. By the definition of $E_A$, $w^*$ is a world in A of minimal rank. Letting B be $\{w^*\}$ in condition (ii), we see that $d(A) \geq d(w^*)$. Again by (ii), $d(w^*) < d(A)$ iff $E_A \cap A$ is empty, which contradicts the definition of $E_A$. Hence $d(A) = d(w^*)$ and we conclude that $d(A) = \min\{rank(w) : w \in A\}$.

Hence we are forced to define the degree of disbelief in a proposition as the minimal degree of disbelief of the worlds in the proposition. But so defined, degrees of disbelief will not help RPMs achieve reversibility of belief change. For in the example given above in which belief in A brought the two rankings $r_1$ and $r_2$ into the same ranking $r_3$, the two initial rankings determine the same degree of belief for A (namely, 1) and for $\neg A$ (namely, 0). But no specification of either of these degrees of disbelief can move $r_3$ back to both $r_1$ and $r_2$.

## 5 DISCUSSION

Researchers on Uncertainty should find the results presented in this paper of interest because they support the view that an adequate account of belief revision (and of reasoning in general) must involve the notion of degrees of belief, so that numerical uncertainty has a prominence in our reasoning that many have been unwilling to grant it. This should not come as a surprise, though, to those aware of the emphasis in recent years on qualitative aspects of probabilistic reasoning (e.g., see Pearl (1988), especially chapter 10).

A stronger result than that proved in this paper would be a *representation theorem* for the Spohn system: a theorem that says a belief revision function satisfies a certain set of axioms iff it coincides with some Spohnian belief revision function. Future research will look at the possibility of obtaining such a result.

### Acknowledgements

I would like to thank Jeff Barnett and Dan Geiger for helpful comments on an earlier version of this paper.